\documentclass{article}
\usepackage{ijcai24}

\usepackage{times}
\usepackage{soul}
\usepackage{url}
\usepackage[hidelinks]{hyperref}
\usepackage[utf8]{inputenc}
\usepackage[small]{caption}
\usepackage{graphicx}
\usepackage{amsmath}
\usepackage{amsthm}
\usepackage{booktabs}
\usepackage{algorithm}
\usepackage{algorithmic}
\usepackage[switch]{lineno}
\usepackage{multirow}
\usepackage{amssymb}
\usepackage{xcolor}
\title{CMMU: A Benchmark for Chinese Multi-modal Multi-type Question Understanding and Reasoning}

\author{
Zheqi He$^1$ \thanks{Equal contribution.}
\and
Xinya Wu$^1$ \footnotemark[1] 
\and
Pengfei Zhou$^{1,2}$ \thanks{This work was done when Zhou was an intern at Beijing Academy of Artificial Intelligence.}
\and
Richeng Xuan$^1$ \and
Guang Liu$^1$ \thanks{Corresponding authors.} \and
Xi Yang $^1$ \footnotemark[3] \and \\
Qiannan Zhu $^3$ \and
Hua Huang  $^3$ \\
\affiliations
$^1$Beijing Academy of Artificial Intelligence\\
$^2$Beijing University of Post and Telecommunication\\
$^3$Beijing Normal University\\
\emails
\{zqhe, yxwu, rcxuan,liuguang,yangxi\}@baai.ac.cn,
zhoupengfei@bupt.edu.cn,
\{zhuqiannan,huahuang\}@bnu.edu.cn
}

\begin{document}

\maketitle

\begin{abstract}
Multi-modal large language models(MLLMs) have achieved remarkable progress and demonstrated powerful knowledge comprehension and reasoning abilities. However, the mastery of domain-specific knowledge, which is essential for evaluating the intelligence of MLLMs, continues to be a challenge. Current multi-modal benchmarks for domain-specific knowledge concentrate on multiple-choice questions and are predominantly available in English, which imposes limitations on the comprehensiveness of the evaluation. To this end, we introduce \textbf{CMMU}, a novel benchmark for multi-modal and multi-type question understanding and reasoning in Chinese. CMMU consists of 3,603 questions in 7 subjects, covering knowledge from primary to high school. The questions can be categorized into 3 types: multiple-choice, multiple-response, and fill-in-the-blank, bringing greater challenges to MLLMs. In addition, we propose an evaluation strategy called \textbf{Positional Error Variance} for assessing multiple-choice questions. The strategy aims to perform a quantitative analysis of position bias. We evaluate seven open-source MLLMs along with GPT4-V, Gemini-Pro, and Qwen-VL-Plus. The results demonstrate that CMMU poses a significant challenge to the recent MLLMs. The data and code are available at \textcolor{blue}{\url{https://github.com/FlagOpen/CMMU}}.

\end{abstract}

\begin{figure*}
    \centering
    \includegraphics[width=1.0\linewidth]{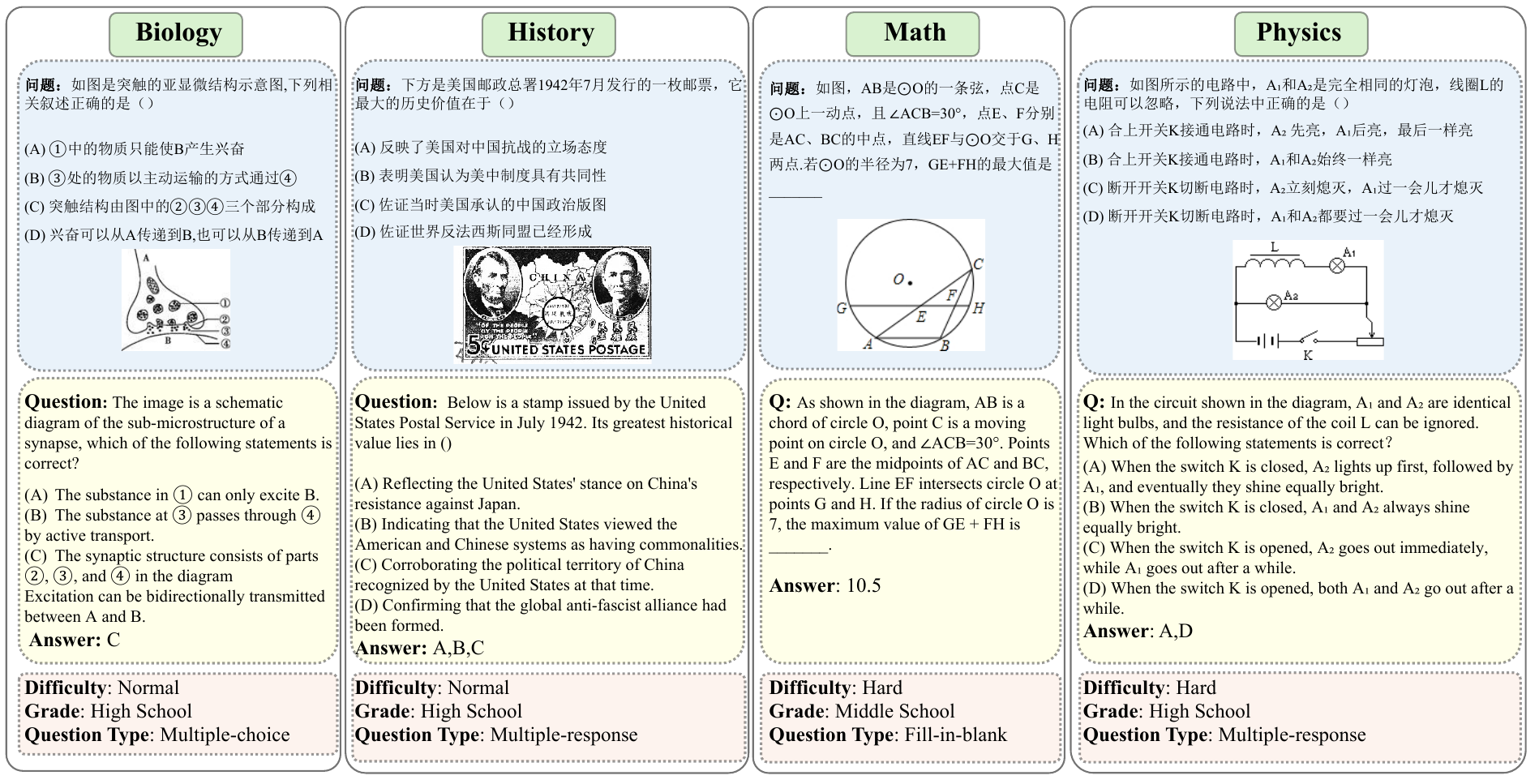}
    \caption{Some examples in CMMU. We provide Chinese examples and their corresponding English translations.}
    \label{fig:show_data}
\end{figure*}

\section{Introduction}
Currently, multi-modal large language models (MLLM) like GPT-4\cite{openai2023gpt}, Gemini\cite{team2023gemini}, LLaVA\cite{liu2023improved} and Qwen-VL \cite{Qwen-VL} have showed powerful abilities in this field of multi-model. At the same time, the ability to evaluate MLLMs more rationally and comprehensively is receiving increasing attention. Researchers have made many efforts to address this problem. Datasets like MMBench \cite{fu2023mme}, MME \cite{fu2023mme}, and SEED-Bench \cite{li2023seed,li2023seed2} evaluate models through a diverse range of questions, ranging from perception to reasoning abilities. However, these datasets primarily access common-scene knowledge more than domain-specific knowledge. The recently introduced GAIA benchmark \cite{mialon2023gaia} focuses on testing fundamental abilities like reasoning, multi-modal processing, and general tool use. However, GAIA also presents certain limitations. It primarily tests tasks that are conceptually simple for humans, which may not fully capture the complex problem-solving capabilities required in some specialized domains. 

% Currently, multi-modal large language models (MLLM) like GPT-4\cite{openai2023gpt}, Gemini\cite{team2023gemini}, LLaVA\cite{liu2023improved} and Qwen-VL \cite{Qwen-VL} have demonstrated powerful abilities in multi-modal tasks. At the same time, how to evaluate MLLMs in a more rational and comprehensive manner is receiving increasing attention, and researchers have made a lot of efforts to address this problem recently. Datasets like MMBench \cite{fu2023mme}, MME \cite{fu2023mme}, and SEED-Bench \cite{li2023seed,li2023seed2} evaluate models through a diverse range of questions, ranging from perception to reasoning abilities. However, these datasets primarily access common-scene knowledge more than domain-specific knowledge. GAIA \cite{mialon2023gaia} focus on constructing complex tasks in real world to examine fundamental abilities like multi-modal understanding, tool using, reasoning and planing. It's a significant challenge for models, even GPT-4 achieves only a 14.6\% average score when utilizing tools, proving excessively challenging for other open-source models.
In addition to the above benchmarks, alternative evaluation datasets containing questions from textbooks and other educational materials are proposed to evaluate domain-specific knowledge, which are inspired by human exams for measuring knowledge levels and selecting talents. For instance, ScienceQA\cite{lu2022learn} is a dataset that evaluates the scientific knowledge of models, while MMMU\cite{yue2023mmmu} assesses university-level knowledge. These two datasets only contain English questions, while some datasets, such as M3Exam \cite{zhang2023m3exam}, turn attention to the multilingual setting. However, the above benchmarks mainly focus on multiple-choice questions, which limits the comprehensiveness of evaluation. Multiple-choice questions cannot evaluate the text generation abilities of the models, as the models only need to choose the correct answer from a few existing options. Meanwhile, the models may obtain correct answers through guessing, which could impact the accuracy of the evaluation. Therefore, there is a need for a diversified and comprehensive benchmark to evaluate the understanding and reasoning abilities of MLLMs.

To bridge the dataset gap, we introduce a novel benchmark, CMMU, for multi-modal and muli-type question understanding and reasoning in Chinese. CMMU encompasses multi-modal content across 7 subjects. Every question requires the model to combine image and text content to generate a comprehensive response. While CMMU shares similarities with datasets like ScienceQA and M3Exam \cite{zhang2023m3exam}, it offers a broader range of question types. Previous datasets only have multiple-choice questions, while CMMU offers a wider variety of question types, including multiple-choice, multiple-response, and fill-in-the-blank questions, which poses a more significant challenge to the comprehension abilities of MLLM. In addition, to mitigate the position bias\cite{zheng2023judging} in LLM and ensure genuine correctness rather than guessing, inspired by CircularEval \cite{liu2023mmbench}, we adopt a \textbf{Positional Error Variance} approach to measure the position bias for multiple-choice question. Specifically, we cycle through the position of options to ensure that the answer can appear at any position with equal probability, which is the same as CircularEval, aim at reducing position bias, minimizing the influence of randomness on correctness. Then we produce \textbf{Positional Error Variance}, a quantitative analysis to measure position bias. We evaluate 11 models using the CMMU benchmark, and the results indicate that CMMU presents a significant challenge to current MLLMs.

% To bridge the dataset gap, we introduce a novel benchmark, CMMU, for multi-modal and muli-type question understanding and reasoning in Chinese. CMMU encompasses multi-modal content across 7 subjects. Every question requires the model to combine image and text content to generate a comprehensive response. While CMMU shares similarities with datasets like ScienceQA and M3Exam \cite{zhang2023m3exam}, it offers a broader range of question types. Previous datasets only have multiple-choice questions, while CMMU offers a wider variety of question types, including multiple-choice, multiple-response, and fill-in-the-blank questions, which poses a more significant challenge to the comprehension abilities of MLLM. In addition, to mitigate the position bias\cite{zheng2023judging} in LLM and ensure genuine correctness rather than guessing, inspired by CircularEval \cite{liu2023mmbench}, we adopt a \textbf{ShiftCheck} approach for multiple-choice question, which includes two steps. Specifically, we cycle through the position of options to ensure that the answer can appear at any position with equal probability, which is the same as CircularEval, aim at reducing position bias, minimizing the influence of randomness on correctness. Then we produce a quantitative analysis method to measure position bias. We evaluate 11 models using the CMMU benchmark, and the results indicate that CMMU presents a significant challenge to current MLLMs. 
\begin{figure}
    \centering
    \includegraphics[width=0.85\linewidth]{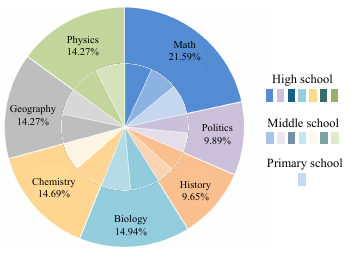}
    \caption{Distribution of questions in subjects and grades.}
    \label{fig:subject_dist}
\end{figure} 
To sum up, our contributions are as follows: 
\begin{itemize}
\item We present a novel benchmark of multi-modal and multi-type questions in Chinese, featuring a wider variety of question types, including multiple-choice, multiple-response, and fill-in-the-blank questions.
\item We evaluate 10 models and analyze their performances in Chinese language proficiency and multi-modal comprehension.
\item We propose Positional Error Variance, which is designed to conduct a quantitative analysis of position bias in MLLMs.
\end{itemize}
\section{Related Work}

\subsection{Multi-modal Benchmarks}
With the development of large language models (LLMs), there is a growing emphasis in research communities on assessing the capabilities of LLM such as HELM\cite{liang2022holistic}, CLEVA \cite{li2023cleva}, especially their multimodal understanding abilities. Datasets like VQAv2 \cite{goyal2017making}, TDIUC \cite{kafle2017analysis}, TextVQA \cite{singh2019towards} and GQA \cite{hudson2019gqa} are used in visual question answering tasks, while COCO \cite{lin2014microsoft}, NoCaps \cite{agrawal2019nocaps}, and Flickr30K \cite{plummer2015flickr30k} are employed in image captioning tasks. Additionally, Visual7w \cite{zhu2016visual7w} and RefCOCO \cite{kazemzadeh2014referitgame} are commonly utilized for visual grounding purposes. With the rapid development of multi-modal large language models, researchers have achieved good results on these datasets. We require more extensive data to evaluate MLLMs, and there have been recent studies evaluating models from various perspectives. LVLM-eHub \cite{xu2023lvlm} collects 47 existing benchmarks and evaluates 6 types of capabilities of MLLMs, however, it does not create any new benchmarks. MME \cite{fu2023mme} comprehensively measures the perception and cognition abilities of models. However, its question types are simplistic, merely requiring \textit{yes} or \textit{no} responses. MMBench \cite{liu2023mmbench} and SEED-Bench \cite{li2023seed,li2023seed2} contain many multiple-choice questions covering various ability dimensions, but these datasets mainly consist of common-sense questions and do not require lots of domain-specific knowledge and complex reasoning. To enhance the evaluation of domain-specific knowledge, ScienceQA \cite{lu2022learn} was introduced. This dataset encompasses a wide range of science topics from elementary and high school curricula. MMMU \cite{yue2023mmmu} is designed to evaluate college-level subject knowledge, questions of CMMU are collected from college exams and textbooks, and many of them require expert-level skills. BenchLMM \cite{cai2023benchlmm} assesses MLMMs in three distinct styles: artistic style, sensor style, and application style. M3Exam \cite{zhang2023m3exam} is a multilingual and multi-modal benchmark designed to evaluate domain knowledge and problem-solving skills, and it spans seven languages. However, less than one-third of the questions include images.
\begin{figure}
    \centering
    \includegraphics[width=1.0\linewidth]{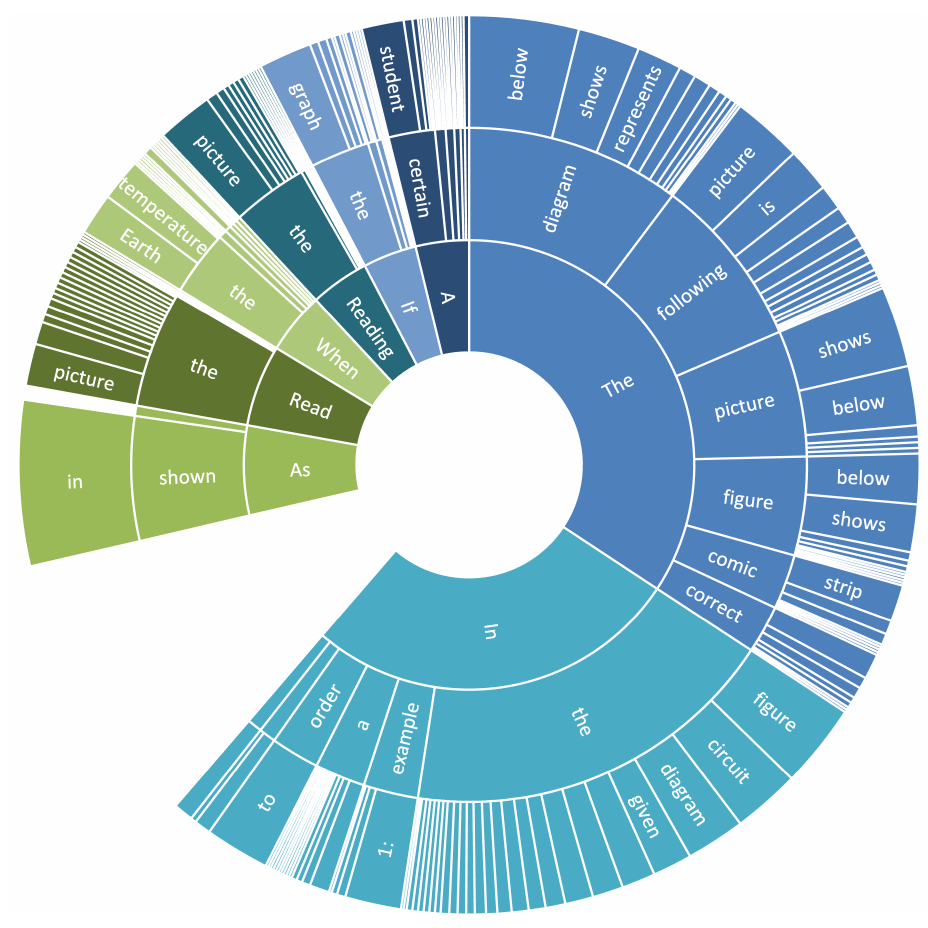}
    \caption{The first three words of questions in CMMU. We have translated it into English.}
    \label{fig:ques-distrubution}
\end{figure}

\subsection{Multi-modal Large Language Models}
Benefiting from the development of large language models(LLM) like GPT \cite{macfarlane2023professional}, LLaMA \cite{touvron2023llama} and Vicuna \cite{chiang2023vicuna}, MLLMs have made significant improvements. Many works integrate additional modal inputs on LLM and bridge the modality gap between vision and language, and the majority of MLLMs possess multilingual capabilities. BLIP-2 \cite{li2023blip} propose Q-Former to align image representation and text representation, InstructBLIP \cite{dai2305instructblip} based on BLIP-2 and propose an instruction tuning framework to improve the instruction following capability. CogVLM \cite{wang2023cogvlm} propose a visual expert module to enable deep alignment of the vision-language features. LLaVA \cite{liu2023visual,liu2023improved}, Emu2 \cite{sun2023generative}, and MiniGPT-4 \cite{zhu2023minigpt} adopt a simple but effective projection scheme to connect image feature into the language space. A modality-adaptive module is introduced by mPLUG-Owl2 \cite{ye2023mplug}, aiming to enhance modality collaboration by projecting visual and linguistic features into a shared space. In this paper, we will provide a comprehensive evaluation of some of these models using the CMMU benchmark and assess their abilities in domain-specific knowledge.

\begin{table}
    \centering
    \renewcommand{\arraystretch}{1.1}
    \begin{tabular}{lc}
    \toprule
    \textbf{Static} &  \textbf{Number} \\
    \hline
    Total Questions & 3,603 \\ 
    Validation:Test & 1,800:1,803 \\ 
    Subjests & 7 \\
    \hline
    Questions with a explanation  & 2,585 (71.75\%)\\
    Difficulties: Normal&  2,885 (80.07\%) \\
    Difficulties: Hard & 718 (19.93\%) \\
    \hline
    Multiple-choice question & 2,710 (75.22\%) \\ 
    Multiple-response question & 254 (7.05\%)\\ 
    Fill-in-the-blank question & 639 (17.74\%)\\ 
    \hspace{0.3cm} *Sub-questions & 1,632\\
    \hline
    Primary school & 250 (6.90\%)\\
    Middle school & 1,697 (47.19\%)  \\
    High school & 1,656  (45.96\%)\\
    \hline
    Average question length & 72.15 \\
    Average sub-question length & 43.91 \\
    Average choice length & 14.47 \\
    Average answer explanation length & 311.67 \\
    \bottomrule
    \end{tabular}
    \caption{Detailed statistics of the CMMU}
    \label{tab:static}
\end{table}
\begin{table}
    \centering
    \small
    \setlength{\tabcolsep}{1pt}
    
    \renewcommand{\arraystretch}{1.1}
    \begin{tabular}{l|cccccc}
    \toprule
         &Q &I &Exp & Question Type &Lang  \\ \hline 
         MMLU &15,687&\checkmark&$\times$&MCQ&en \\
         MMBench &2,974&\checkmark&$\times$&MCQ&en,ch\\
         SCI.QA &21,208&\checkmark&\checkmark&MCQ&en\\
         M3Exam &12,317&\checkmark&$\times$&MCQ&multilingual\\
         M3KE &20,477&$\times$&$\times$&MCQ&zh \\
         MME & 2,374 &\checkmark &$\times$& True or False &en \\
         MMMU &11,500&\checkmark&\checkmark &MCQ&en  \\
         CMMU &3,603&\checkmark &\checkmark &MCQ, FBQ, MRQ&zh \\
         
    \bottomrule
    \end{tabular}
    \caption{Compare CMMU with existing datasets. Q means quantity, I means image, Exp means explanation of the answer, Lang means language.}
    \label{tab:comparisons}
\end{table}
\begin{table*}
    \centering
    \renewcommand{\arraystretch}{1.05}
    \setlength{\tabcolsep}{2.5pt}
    \begin{tabular}{l|cc|ccc|ccc|ccc|ccc}
    \toprule
         & {Val} & {Test} & \multicolumn{3}{c|}{Val-Normal}   & \multicolumn{3}{c|}{Val-Hard} & \multicolumn{3}{c|}{Test-Normal}   & \multicolumn{3}{c}{Test-Hard}\\
         &{Avg.}  &{Avg.}  & {MCQ}  &{MRQ} &{FBQ}  & {MCQ}  &{MRQ} &{FBQ}  & {MCQ}  &{MRQ} &{FBQ}  & {MCQ}  &{MRQ} &{FBQ}\\
         \hline
         InstructBLIP-13b & 0.39 & 0.48 &0.0 & 0.0 & 0.79 & 0.0 & 0.0 & 1.67 & 0.08 & 0.0 & 1.7 & 0.0 & 1.05 & 0.0\\
         CogVLM-7b& 5.55 & 4.9 &5.98 & 0.0 & 6.9 & 2.0 & 2.13 & 5.0 & 5.89 & 0.0 & 5.1 & 0.67 & 0.0 & 4.73\\
         ShareGPT4V-7b& 7.95 &  7.63 &8.71 & 0.0 & 9.27 & 7.33 & 1.06 & 6.0 & 8.38 & 0.0 & 10.4 & 2.67 & 0.0 & 5.41\\  
         mPLUG-Owl2-7b&  8.69 & 8.58 &10.62 & 3.03 & 8.28 & 6.67 & 1.06 & 5.67 & 9.63 & 0.0 & 11.15 & 5.33 & 1.05 & 4.73\\
         LLava-1.5-13b& 11.36 &11.96&12.7 & 0.0 & 12.62 & 8.67 & 1.06 & 9.67 & 13.03 & 3.12 & 14.93 & 6.67 & 0.0 & 9.8\\
         Qwen-VL-Chat-7b & 11.71 &12.14 &9.71 & 3.03 & 17.36 & 3.33 & 1.06 & 18.67 & 10.62 & 0.0 & 21.36 & 0.67 & 1.05 & 12.5\\   
         Intern-XComposer-7b & 17.87 & 18.42 &22.49 & 3.03 & 16.96 & 8.67 & 4.26 & 11.33 & 22.16 & 12.5 & 20.04 & 7.33 & 1.05 & 12.16\\
         \hline
         Gemini-Pro & 21.58 & 22.5 &18.42 & \textbf{24.24} & 33.53 & 5.33 & 17.02 & 23.33 & 20.83 & 21.87 & 31.95 & 4.67 & 11.58 & 25.0\\
         Qwen-VL-Plus& 27.51  & 27.73 &26.33 & 12.5 & 34.98 & \textbf{19.46} & 14.89 & 29.19 & 28.31 & 28.12 & 31.19 & \textbf{22.82} & 10.53 & \textbf{27.12}\\
         GPT-4V &  \textbf{30.19} & \textbf{30.91} &\textbf{30.54} & 21.21 & \textbf{35.31} & 14.67 & \textbf{23.4} & \textbf{31.0} & \textbf{32.86} & \textbf{37.5} & \textbf{37.81} & 12.67 & \textbf{16.84} & 23.65\\
    \bottomrule
    \end{tabular}
    \caption{The accuracy of comparing models on different question types and difficulty levels. We report the results of the models on the validation and test sets. }
    \label{tab:overall_results}
\end{table*}
\section{The CMMU Benchmark}

CMMU is a novel multi-modal benchmark designed to evaluate domain-specific knowledge across seven foundational subjects: math, biology, physics, chemistry, geography, politics, and history. We collect questions encompassing both text and images sourced from diverse exams. CMMU covers questions from primary school to high school, providing a comprehensive evaluation of the abilities of models across various grades.

Previous benchmarks, such as ScienceQA and MMMU, only have multiple-choice questions. In contrast, our CMMU benchmark contains 3 types of questions: 
\begin{itemize}
\item Multiple-choice question (MCQ): Each question presents 3 or 4 options, with only one correct answer. 
\item Multiple-response question (MRQ): Each question includes 4 options, and the number of correct answers can range from 1 to 4. 
\item Fill-in-the-blank question (FBQ): The question is to fill in the blanks with the correct answers to complete the sentence or passage. 
\end{itemize}
In addition to providing the correct answer, CMMU also provides the explanations of the answers about MCQ and MRQ.
% This diverse question structure closely emulates real exams while also enabling automated evaluation.
\begin{figure}
    \centering
    \includegraphics[width=0.9\linewidth]{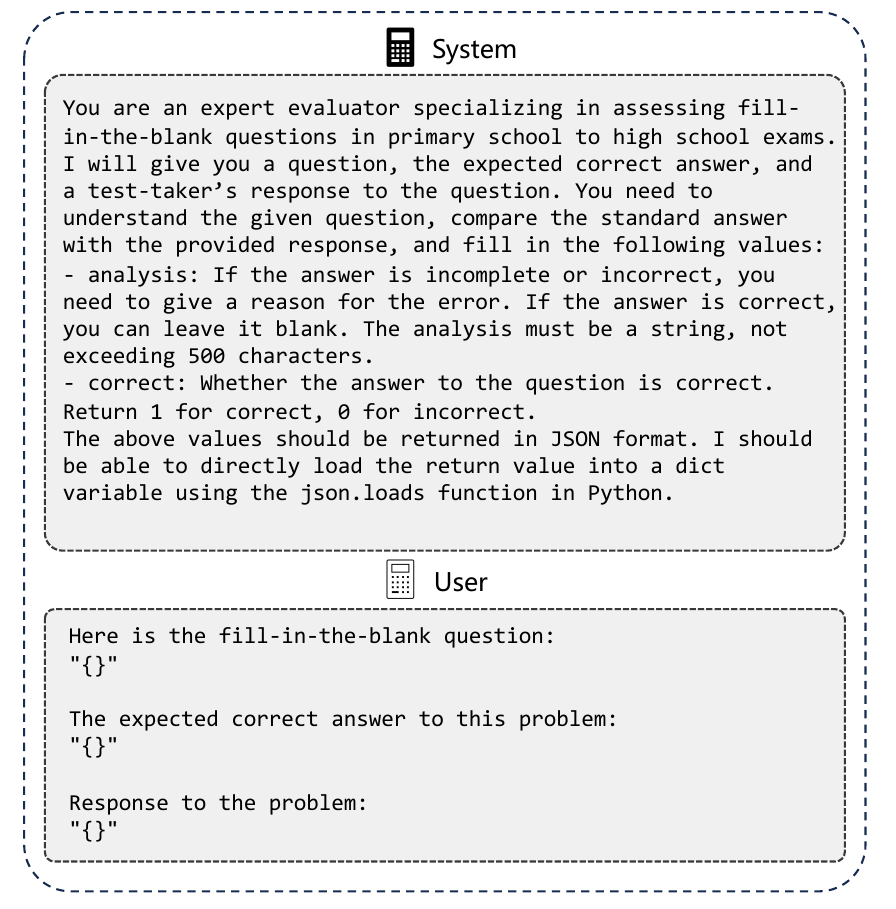}
    \caption{Prompt template used in fill-in-the-blank questions}
    \label{fig:eval-prompt}
\end{figure}

\begin{figure}
    \centering
    \includegraphics[width=1\linewidth]{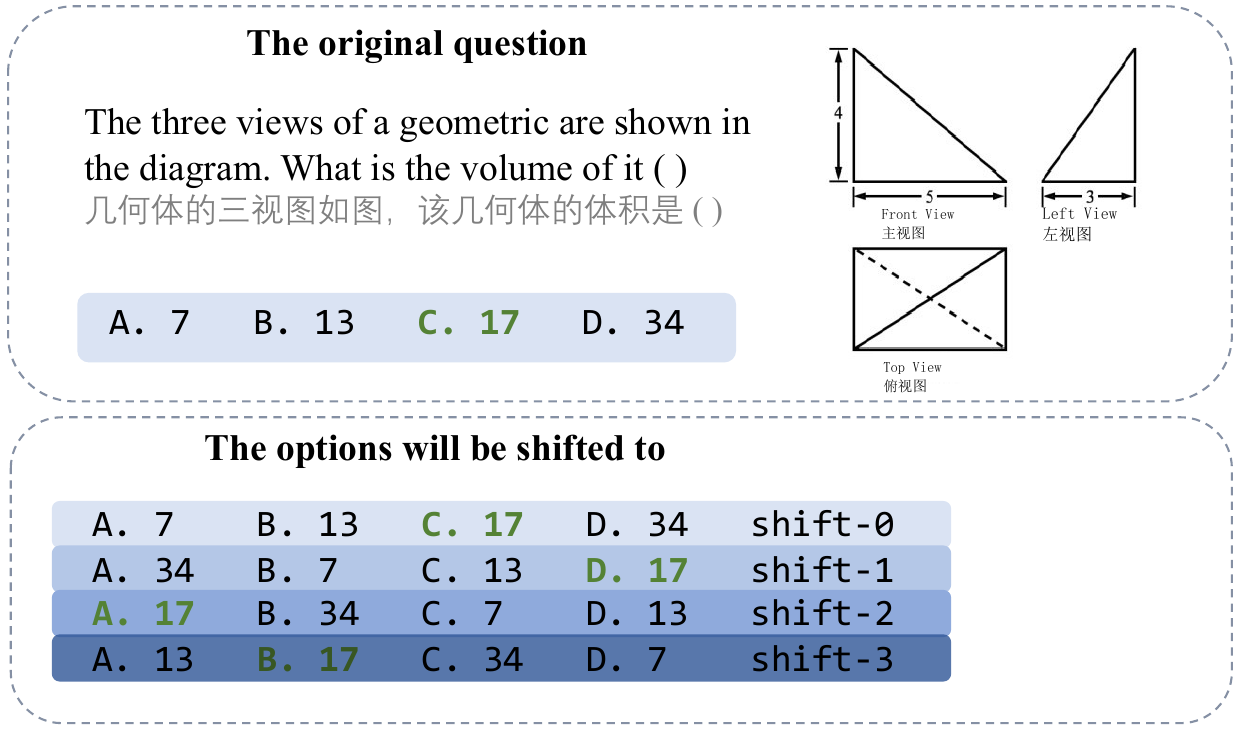}
    \caption{A demonstration of the CircularEval.}
    \label{fig:shift_test}
\end{figure}

\begin{figure*}
    \centering
    \includegraphics[width=1.0\linewidth]{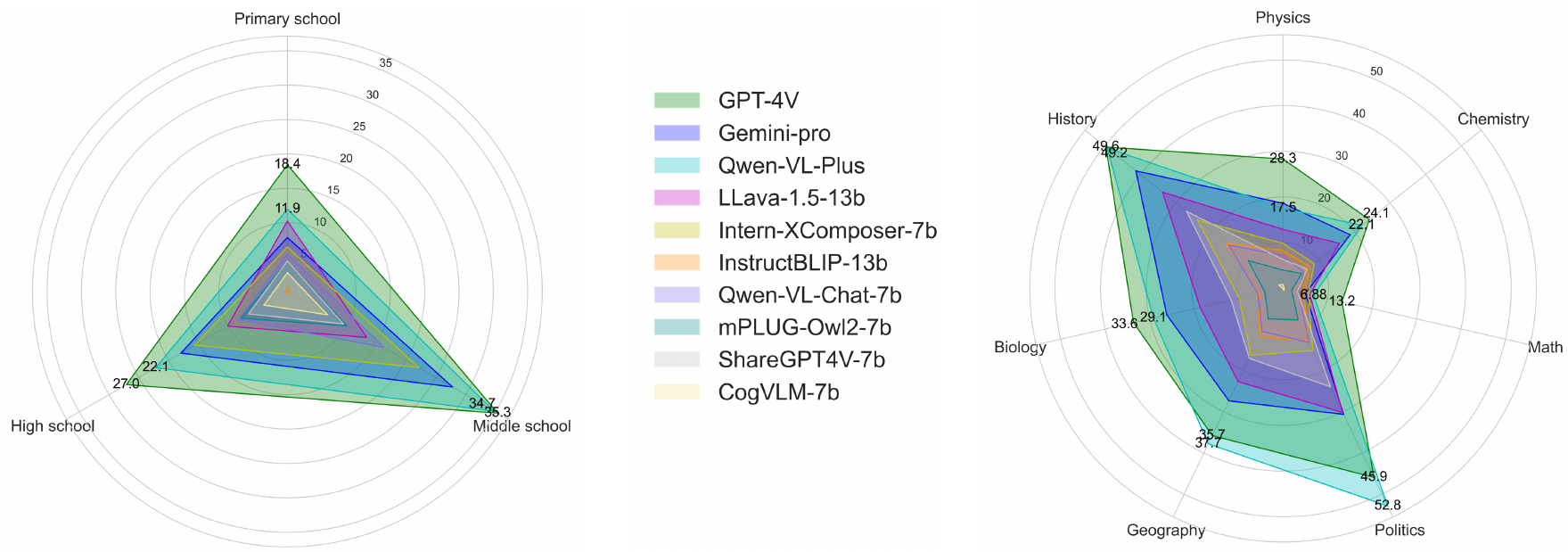}
    \caption{Overall results on the accuracy of different models in subjects and grades.}
    \label{fig:radar}
\end{figure*}
\subsection{Data Pre-process}
\textbf{Data Collection and Processing.} We extract text and images from the original PDF and convert them into JSON format. In addition, we transform all formulas, including mathematical and chemical ones, into LaTeX format. For fill-in-the-blank questions, if there are many sub-questions within one question, we will split them into a sub-question list, attempting to have only one blank to fill in each sub-question, except for some indivisible questions. In the end, we decompose 639 fill-in-the-blank questions into 1,632 sub-questions, with 83\% of them requiring only one blank to be filled.

\noindent
\textbf{Data Cleaning.} We manually review the questions, filtering out images that are blurry, low-quality, or have a resolution less than 50$\times$50 dpi, eliminating questions that are incorrectly parsed, and correcting mistakes made during the automatic conversion to LaTeX. Furthermore, experienced teachers consider the depth of knowledge and the complexity of question-solving methods to categorize each question into two levels: normal and hard based on their experience and subjective judgment.

\subsection{Data Distribution}
CMMU has a total of 3603 questions, divided into validation set and test set, with 1,800 and 1,803 questions respectively. The validation set is open source to the community. As shown in Figure \ref{fig:subject_dist}, the benchmark contains multi-modal content from middle and high school across 7 types of subjects, while primary school only contains math. The ratio of normal and hard questions are 8:2. Over 70\% of the questions have detailed answer explanations, with an average length of 311 characters for each analysis. The detailed statistics are shown in Table \ref{tab:static}. We translate the original questions into English and analyze the distribution of their first three words. As shown in Figure \ref{fig:ques-distrubution}, the questions have diverse formats and are relevant to images or diagrams. 
% The questions are categorized into three grades: primary school, middle school, and high school.

\subsection{Comparisons with Existing Datasets}
Table \ref{tab:comparisons} shows the comparisons with existing datasets. We compare the benchmarks from five dimensions: quantity, with or without images, with or without explanation, question type and language. It can be seen that CMMU is the first benchmark for multi-modal and muli-type question understanding and reasoning in Chinese.

\section{Evaluation}

\label{shiftcheck}
The evaluation of multiple-choice questions confronts two challenges: First, considering the particularity of the formats of multiple-choice questions, when the model correctly answers a question there is an uncertainty about whether the model has truly mastered the relevant knowledge or it just guesses the correct answer. When a model chooses answers through guessing, there will be positional bias, which means a LLM will prefer the answer in a certain position. Position bias is an issue that appears in many LLMs and MLLMs, however, existing methods have not quantitatively measured the extent of the position bias.

To address the above problems, we propose Positional Error Variance. Firstly, following the CircularEval \cite{liu2023mmbench}, we cyclically change the positions of the options and let the model answer questions. Subsequently, we calculate metrics to quantify position bias. We will describe the whole process in detail.

\subsection{CircularEval} For a multiple-choice question with $k$ options, we perform a right circular shift on the options. For example, if the original order of the options is $ABCD$, then after one shift, the order will change to $DABC$. A detailed example is provided in Figure \ref{fig:shift_test}. Given a question $Q$ with $k$ possible options, we generate $k$ distinct shifted-option questions, denoted as $Q_{i}$, \(i \in [0, k]\). Each $Q_{i}$ is then evaluated by the MLLMs to generate the corresponding answers $A_{i}$. We consider the model to have sufficient knowledge to answer the question $Q$ only if all of $A_{i}$ are correct, in which case the accuracy score of $Q$ is 1, otherwise it is 0.
\begin{table*}
    \centering
    \setlength{\tabcolsep}{2pt}
    \renewcommand{\arraystretch}{1.05}
    \begin{tabular}{l|ccccccc|ccc}
    \toprule
         &Physics &Chemistry &Math &Politics &Geography &Biology &History &Primary &Middle &High\\
         \hline
         InstructBLIP-13b &0.92 & 0.26 & 0.24 & 0.48 & 0.0 & 0.12 & 1.17 &0.68 & 0.45 & 0.29\\
         CogVLM-7b &4.0 & 5.23 & 1.93 & 7.66 & 7.41 & 4.05 & 9.79 &2.73 & 6.74 & 3.95\\
         ShareGPT4V-7b &7.23 & 5.75 & 3.49 & 13.16 & 10.47 & 5.64 & 15.15 &4.44 & 9.44 & 6.5 \\
         Qwen-VL-Chat-7b &6.31 & 6.8 & 3.25 & 23.92 & 17.01 & 11.66 & 27.04 &5.46 & 16.27 & 8.18 \\
         mPLUG-Owl2-7b &8.31 & 7.58 & 4.69 & 12.68 & 11.77 & 5.4 & 15.85 &4.44 & 9.98 & 7.8\\
         LLava-1.5-13b &9.85 & 8.76 & 5.66 & 15.07 & 16.28 & 9.82 & 24.01 &10.24 & 13.35 & 10.06  \\
         Intern-XComposer-7b &12.92 & 15.82 & 6.02 & 30.14 & 22.67 & 18.65 & 33.8 & 6.48 & 22.07 & 15.59 \\
         \hline
         Qwen-VL-Plus &17.57 & 22.12 & 6.88 & \textbf{52.88} & \textbf{37.76} & 29.15 & 49.29 &11.99 & 34.79 & 22.14\\
         Gemini-Pro &18.62 & 18.82 & 5.05 & 30.62 & 27.33 & 26.13 & 41.26 & 7.85 & 27.78 & 17.9 \\
         GPT-4V &\textbf{28.31} & \textbf{24.18} & \textbf{13.24} & 45.93 & 35.76 & \textbf{33.62} & \textbf{49.65} &\textbf{18.43} & \textbf{35.37} & \textbf{27.09}\\
    \bottomrule
    \end{tabular}
    \caption{Detailed statistics of different models in subjects and grades. We average the accuracy of different difficulty questions and report the average values on the test and validation sets.}
    \label{tab:subject_results}
\end{table*}

\subsection{Positional Error Variance} Conceptually, an unbiased model assigns equal probability to each option. Under the shifted-option setting, if the probability of each option is not equal, it indicates that the model has a bias towards a certain option. Considering this, we define the $BiasRate$ as follows: 

Questions that are completely answered correctly in CircularEval do not reflect position bias, so we just focus on the incorrectly answered questions. If there are $M$ incorrectly answered questions with $n$ options for each, there will be a total of $m * n$ answers combination. We count the occurrences $S_o$ of each option $o$ and then calculate the probability $P_o = \frac{S_o}{m * n}, o \in \{A, B, C,...\}
$. And then we define the BiasRate as the variance of $P$, the formula is $BiasRate = \sigma^2(P)$. The larger the $BiasRate$, the greater the positional bias of the model.

\subsection{Evaluations on Different Question Types}
To avoid the impact of the analysis process of model outputs on the evaluation, we filter the answers by retaining only the last line of the answer. For multiple-choice and multiple-response questions, we extract option letters from the responses of the models. After that, we apply different strategies to evaluate the three types of questions.

% We find that the instruction-following ability of the models is limited, and it may output excessive information, such as analytical processes, when answering questions. Prior to evaluation, we perform an initial cleanup of the answers. Specifically, we retain only the last line of the answer. For multiple-choice and multiple-response questions, we extract option letters from the model's response. After that, we apply different strategies to evaluate the three types of questions.

\textbf{Evaluation on Multiple-choice Question}: We adopt CircularEval and Positional Error Variance in section \ref{shiftcheck}, which allows us to analyze both the accuracy and the $BiasRate$.

\textbf{Evaluation on Multiple-response Question}: This question type may have more than one correct option. We consider the correctness only when all the chosen options are correct, excluding any incorrect choices.

\textbf{Evaluation on Fill-in-the-blank Question}: The answers to fill-in-the-blank questions may not be unique and responses with similar meanings to the groundtruth can also be considered correct. Hence, we utilize GPT-4 to judge the answer, providing a binary score of 0 or 1 to determine correctness. Further details about the evaluation prompts are in Figure \ref{fig:eval-prompt}.

\section{Experiments}

\subsection{Models}
We evaluate the performance of various MLLMs, including both closed-source and open-source models. The closed-source models are evaluated by using their official API, while open-source models are evaluated by running inferences on NVIDIA A100 GPUs. For the closed-source models, we select state-of-the-art models like GPT-4V, Gemini-Pro. We also choose Qwen-VL-Plus, which performs well on Chinese datasets. For the open-source models, model sizes vary from 7b to 13b, including LLava-1.5-13b, CogVLM-7b, InstructBLIP-13b, Qwen-VL-Chat-7b, Intern-XComposer-7b, mPLUG-Owl2-7b and ShareGPT4V-7b.

\subsection{Prompts and Settings}
All models are tested in zero-shot settings as we only specify the output format in prompts. Each type of question has its own prompt template, and we utilize the same prompt template for all models. The prompt\footnote{In the experiments, we use the Chinese version and we translate it into English for reading.} of MCQ is ``Answer with the option's letter from the given choices directly'', the prompt of MRQ is ``Please directly provide the letters of the correct options. There may be more than one correct option.'' the prompt of FBQ is ``Complete each blank with a single word or phrase. If there is more than one blank, split answers with a semicolon (;)''.

Parameters are configured with distinct settings for each question type. For MCQ and MRQ, the temperature is set to 0, and the max new token is set to 10. For FBQ, the temperature is set to 0.2, and the max new token is set to 128. 

\begin{figure}
    \centering
    \includegraphics[width=1\linewidth]{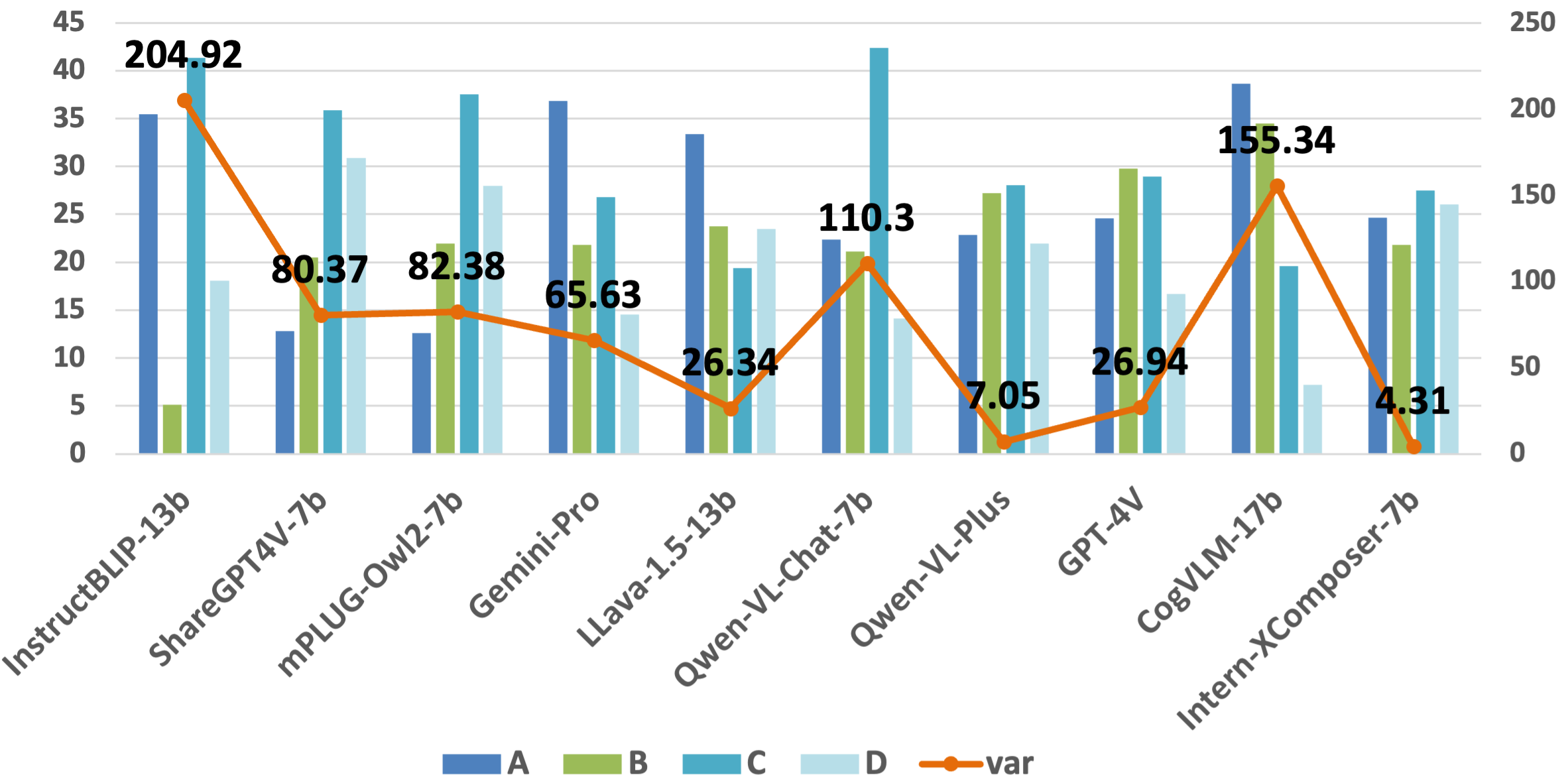}
    \caption{Option distribution and bias rate for different models}
    \label{fig:bias_rate}
\end{figure}

\subsection{Results}

\begin{figure*}
    \centering
    \includegraphics[width=0.95\linewidth]{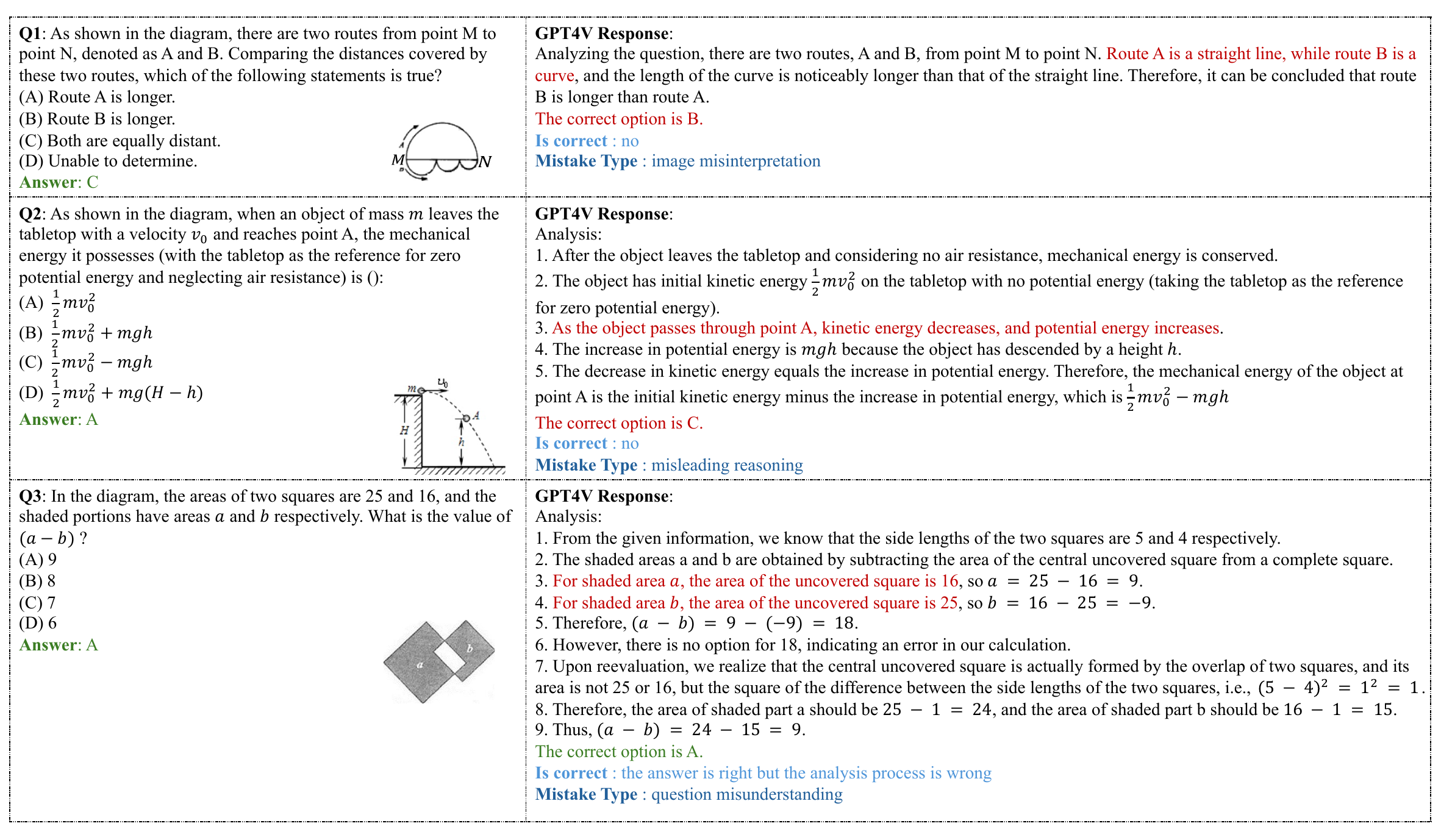}
    \caption{Case study of GPT-4V. Green represents the correct answers to questions, red shows where errors occur in the reasoning process of the model, and blue provides the types of errors. The text is translated into English for reading. }
    \label{fig:case-study}
\end{figure*}
\subsubsection{Results on Different Subjects and Grades}
The overall results are shown in Table \ref{tab:overall_results}. GPT-4V achieves an accuracy of 30.19\% and 30.91\% on the validation and test sets, respectively, reaching the highest level among all models. Moreover, all three closed-source models perform better than all open-source models. From the perspective of question types, most models show lower accuracy in FBQ and MRQ compared to MCQ. Specifically, 7 models achieve an accuracy of less than 10\% on the Val-hard set of MRQ, and 5 models achieve an accuracy of less than 10\% on the Val-hard set of FBQ, highlighting the difficulty and challenge of FBQ and MRQ. 

\subsubsection{Results on Different Question Types}
Figure \ref{fig:radar} shows the overall results of different models in subjects and grades, and the detailed statistics are shown in Table \ref{tab:subject_results}. We can see that there is a significant subject bias in all models. Subjects such as politics and history rely on the knowledge reservoir have higher accuracy than subjects such as physics, math which require computation and reasoning. When comparing different grades, the results show that MLLMs generally perform better on middle school questions than high school ones, suggesting that more complex knowledge presents a more significant challenge for these models.
\subsubsection{Position Bias Analysis}
We employ the \textbf{Positional Error Variance} for quantitative analysis of position bias. As shown in Figure \ref{fig:bias_rate}, most models have a positional preference for one or two specific options. An interesting finding is that, although these positional preferences are inconsistent across models, none of them choose Option $D$ as their most preferred choice. By analyzing the $BiasRate$, we find that superior models, such as GPT-4V, tend to have a relatively lower $BiasRate$.

\subsubsection{Case Study with CoT Prompts}

To further analyze the performance of models using Chain of Thought(CoT), we change the prompt of MCQ to ``Please analyze the question step by step and eventually provide a single correct option letter. (This is a multiple-choice question.)'' Then, we choose GPT-4V, which has a strong ability in instruction-following, to answer 500 randomly selected MCQs. We identify three common mistake types in the model outputs: image misunderstanding, misleading reasoning, and question misunderstanding, with proportions of 27.48\%, 35.41\%, and 13.03\%, respectively. Cases in Figure \ref{fig:case-study} show the above common mistakes respectively: In Question 1, the model fails to identify the route A in the image correctly. In Question 2, the model thinks that the kinetic energy during free fall is transformed into gravitational potential energy, leading to an incorrect reasoning result. In Question 3, the model misunderstands the question and produces a hallucination that the overlap area is 1. Although it guesses the answer correctly, it cannot pass the CircularEval. All bad cases demonstrate that even one of the most advanced MLLMs cannot perfectly solve questions of CMMU, highlighting both the potential and challenges of the benchmark.

\section{Conclusion and Future Work}
% \section*{Acknowledgments}
In conclusion, our work introduces a novel benchmark named CMMU to evaluate the multi-modal and multi-type question understanding and reasoning abilities of MLLMs in Chinese. Unlike existing benchmarks focusing on multiple-choice questions, CMMU offers a more comprehensive evaluation by incorporating a broader question type, including MCQ, MRQ, and FBQ. We also propose Positional Error Variance to quantify the position bias of the model. The evaluation results contribute to a deeper understanding of current MLLMs in the context of diverse and complex question formats. In future work, we will consider enriching the problem types of CMMU and expanding the number of problems to increase the challenge of the benchmark further.

% \section*{Acknowledgments}
% This work is supported by National Key R\&D Program of China (2022ZD0116306). Thanks to Duo Zheng and Ning Su for their valuable suggestions and contributions to this paper.
\bibliographystyle{named}
\bibliography{ijcai24}

\end{document}